\documentclass[runningheads]{llncs}
\usepackage{graphicx}
\usepackage[left=1.5in, right=1.5in, top=1.5in, bottom=1.5in]{geometry}
\usepackage{amsmath}
\usepackage{amsfonts}
\usepackage{amssymb}
\usepackage{hyperref}
\usepackage{booktabs}
\usepackage{multirow}
\usepackage[table,xcdraw]{xcolor}
\usepackage[numbers]{natbib}
\usepackage{float}

\begin{document}
\title{AutoPET Challenge III: Testing the Robustness of Generalized Dice Focal Loss trained 3D Residual UNet for FDG and PSMA Lesion Segmentation from Whole-Body PET/CT Images}
%
%
\author{Shadab Ahamed\inst{1, 2}}
\institute{University of British Columbia, Vancouver, BC, Canada \and
BC Cancer Research Institute, Vancouver, BC, Canada}
\maketitle              
\begin{abstract}

Automated segmentation of cancerous lesions in PET/CT scans is a crucial first step in quantitative image analysis. However, training deep learning models for segmentation with high accuracy is particularly challenging due to the variations in lesion size, shape, and radiotracer uptake. These lesions can appear in different parts of the body, often near healthy organs that also exhibit considerable uptake, making the task even more complex. As a result, creating an effective segmentation model for routine PET/CT image analysis is challenging. In this study, we utilized a 3D Residual UNet model and employed the Generalized Dice Focal Loss function to train the model on the AutoPET Challenge 2024 dataset. We conducted a 5-fold cross-validation and used an average ensembling technique using the models from the five folds. In the preliminary test phase for Task-1, the average ensemble achieved a mean Dice Similarity Coefficient (DSC) of 0.6687, mean false negative volume (FNV) of 10.9522 ml and mean false positive volume (FPV) 2.9684 ml. More details about the algorithm can be found on our GitHub repository: https://github.com/ahxmeds/autosegnet2024.git. The training code has been shared via the repository: https://github.com/ahxmeds/autopet2024.git.

\end{abstract}

\section{Introduction}
\label{sec:introduction}
PET/CT imaging is widely regarded as the gold standard in the management of cancer patients, providing accurate diagnoses, detailed staging, and critical insights for assessing treatment response \cite{pet_importance}. Traditionally, PET/CT images are evaluated qualitatively by radiologists or nuclear medicine specialists. However, this method is subject to potential errors due to variations in interpretation between different experts. By incorporating quantitative analysis into the assessment of PET images, clinical decision-making could become more precise, leading to better prognostic, diagnostic, and staging outcomes for patients undergoing a range of therapeutic treatments \cite{tmtv, suvmean_tmtv_tlg}.\\

Manual segmentation of lesions from PET/CT images by experts is a labor-intensive process and is prone to variability both within and between observers  \cite{FOSTER201476,ahamed2023comprehensive}. This variability underscores the need for automation to make quantitative PET analysis more feasible in everyday clinical settings. Traditional automated methods, such as those based on thresholding, often fail to detect lesions with low radiotracer uptake and can generate false positives in areas with high physiological uptake, like the brain or bladder. Deep learning provides a promising alternative, offering the potential to streamline lesion segmentation, minimize variability, improve efficiency, and enhance the detection of difficult lesions, thereby offering valuable support to healthcare providers \cite{Wang2021, Tang2019, Hollon2020}.

Despite the progress made in deep learning-based segmentation, accurately segmenting lesions in PET/CT images remains a complex challenge. A major obstacle lies in the limited availability of large, well-annotated, and publicly accessible datasets. Most deep learning models are trained using relatively small, privately-owned datasets, which limits their ability to generalize and slows their integration into routine clinical use. Initiatives like the AutoPET segmentation challenge (MICCAI) have been instrumental in addressing this issue by providing large, publicly accessible PET/CT datasets \cite{Gatidis2022, autopet2022_paper}. These challenges represent a pivotal step forward in the development of high-precision models that can meet the rigorous standards required for clinical adoption. The AutoPET dataset, in particular, is noteworthy for its breadth and diversity, featuring patients with various cancer types such as lymphoma, lung cancer, and melanoma, along with a group of negative controls. Unlike the previous versions of the challenge which consisted of only $^{18}$F-fluorodeoxyglucose (FDG) PET images in the training set, the AutoPET-III (2024) expanded the training set to include images from both FDG and prostate-specific membrane antigen (PSMA) PET images. The latter consisted of male patients presenting prostate cancer as well as a small number of negative controls. This wide-ranging composition holds the potential to improve the dataset's representativeness and enhances its potential impact on medical applications.

In this work, we trained a 3D Residual UNet using the provided training dataset and participated in the Task 1 of the challenge. Testing was first performed by submitting the trained algorithm to the challenge preliminary test phase which consisted of 5 cases. The algorithm was then submitted to the final test phase which consisted of 200 cases consisting of both FDG and PSMA PET/CT images for final evaluation.

\section{Materials and methods}
\label{sec:materials_and_methods}
\subsection{Data and data split}
\label{subsec:data_and_datasplit}
The training data consisted of 1014 FDG cases from 900 patients and 597 PSMA cases from 378 patients. The former set consisted of patients presenting lymphoma, lung cancer, melanoma, and negative control patients, while the latter consisted of prostate cancer and negative controls. The data was randomly split into 5 folds (we used the same 5 fold split provided by the challenge organizers from the nnUNet-baseline). No other dataset (public or private) was used in this work.   

\subsection{Preprocessing and data augmentation}
\label{subsec:preprocessing_and_data_augmentation}
The CT images were first downsampled to match the coordinates of their corresponding PET images. The PET intensity values in units of Bq/ml were decay-corrected and converted to SUV. During training, we employed a series of non-randomized and randomized transforms to augment the input to the network. The non-
randomized transforms included (i) clipping CT intensities in
the range of [-1024, 1024] HU (ii) min-max normalization of clipped CT intensity to
span the interval [0, 1], (iii) cropping the region outside the body in PET, CT, and mask images using a 3D bounding box, and (iv) resampling the PET, CT, and mask images to an isotropic voxel spacing of (2.0 mm, 2.0 mm, 2.0 mm) via bilinear interpolation for PET and CT images and nearest-neighbor interpolation for mask images. \\

On the other hand, the randomized transforms were called at the start of every epoch. These included (i) random spatial cropping of cubic patches of dimensions $(N,N,N) = (128, 128, 128)$ from the images, (ii) 3D translations in the range (0, 10) voxels along all three directions, (iii) axial rotations by angle $\theta \in (-\pi/12, \pi/12)$, (iv) random scaling by a factor of 1.1 in all three directions, (v) 3D elastic deformations using a Gaussian kernel with standard deviation and offsets on the grid uniformly sampled from (0, 1), (vi) Gamma correction with $\gamma \in (0.7, 1.5)$, and (vii) addition of random Gaussian noise with $\mu = 0$ and $\sigma=1$. Finally, the augmented PET and CT patches were concatenated along the channel dimension to construct the final input for the network.

\subsection{Network}
\label{subsec:network}
We used a 3D Residual UNet \cite{resunet, ahamed2023generalized}, adapted from the MONAI library \cite{monai_paper}. The network consisted of 2 input channels, 2 output channels, and 5 layers of encoder and decoder (with 2 residual units per block) paths with skip-connections. The data in the encoder
was downsampled using strided convolutions, while the decoder unsampled using transpose strided convolutions. The number of channels in the encoder part from the top-most layer to the bottleneck were 32, 64, 128, 256, and 512. PReLU was used as the activation function within the network. The network consisted of 19,223,525 trainable parameters.

\subsection{Loss function, optimizer, and scheduler}
\label{subsec:loss_function_optimizer_and_scheduler}
We employed the binary Generalized Dice Focal Loss $\mathcal{L}_{\text{GDFL}} = \mathcal{L}_\text{GDL} + \mathcal{L}_\text{FL}$, where $\mathcal{L}_\text{GDL}$ is the  Generalized Dice Loss \cite{gendiceloss} and $\mathcal{L}_\text{FL}$ is the Focal Loss \cite{focalloss}. The Generalized Dice Loss $\mathcal{L}_\text{GDL}$ is given by,
\begin{equation}
 \mathcal{L}_\text{GDL} = 1 - \frac{1}{n_b} \sum_{i=1}^{n_b} \frac{\sum_{l=0}^{1} w_{il} \sum_{j=1}^{N^3} p_{ilj} g_{ilj}  + \epsilon}{\sum_{l=0}^{1} w_{il} \sum_{j=1}^{N^3}(p_{ilj} + g_{ilj})  + \eta}
\end{equation}
where $p_{ilj}$ and $g_{ilj}$ are values of the $j^{th}$ voxel of the $i^{th}$ cropped patch of the predicted and ground truth segmentation masks with class $l \in \{0, 1\}$ respectively in a mini-batch size $n_b$ of the cropped patches and $N^3$ represents the total number of voxels in the cropped cubic patch of size $(N, N, N)$, where $N=128$. Here, $w_{il} = 1/(\sum_{j=1}^{N^3}g_{ilj})^2$ represents the weight given to class $l$ for the $i^{\text{th}}$ cropped patch in the batch. The mini-batch size was set to $n_b=4$. Small constants $\epsilon = \eta = 10^{-5}$ were added to the numerator and denominator, respectively to ensure numerical stability during training. The Focal Loss $\mathcal{L}_\text{FL}$ is given by,
\begin{equation}
    \mathcal{L}_\text{FL} = -\frac{1}{n_b}\sum_{i=1}^{n_b} \sum_{l=0}^{1} \sum_{j=1}^{N^3} v_l (1 - \sigma(p_{ilj}))^\gamma g_{ilj} \text{log}(\sigma(p_{ilj}))
\end{equation}
where, $v_0 = 1$ and $v_1 = 100$ are the focal weights of the two classes, $\sigma(x) = 1/(1 + \text{exp}(-x))$ is the sigmoid function, and $\gamma = 2$ is the focal loss parameter that suppresses the loss for the class that is easy to classify ($l = 0$ or background class in our case).\\

$\mathcal{L}_{\text{GDFL}}$ was optimized using the Adam optimizer. Cosine annealing scheduler was used to decrease the learning rate from $1 \times 10^{-3}$ to $0$ in 400 epochs. The loss for an epoch was computed by averaging the $\mathcal{L}_{\text{GDFL}}$ over all batches. The model with the highest mean Dice similarity coefficient (DSC) on the validation fold $f$ was chosen for further evaluation, for all $f \in \{0, 1, 2, 3, 4\}$.
\subsection{Inference and postprocessing}
\label{subsec:inference_and_postprocessing}
For the images in the validation set, we employed only the non-randomized transforms. The prediction was made directly on the 2-channel (PET and CT) whole-body images using a sliding-window technique with a window of dimensions $(192, 192, 192)$ and overlap=0.5. For final testing, the outputs of the 5 best models (obtained from 5-folds training) were ensembled via average ensembling to generate the output mask. The final output masks were resampled to the coordinates of the original ground truth masks for computing the evaluation metrics.

\subsection{Evaluation metrics}
The challenge employed three evaluation metrics, namely the mean DSC, mean false positive volume (FPV) and mean false negative volume (FNV). For a foreground ground truth mask $G$ containing $L_g$ disconnected foreground segments (or lesions) $\{G_1, G_2, ..., G_{L_g}\}$ and the corresponding predicted foreground mask $P$ with $L_p$ disconnected foreground segments $\{P_1, P_2, ..., P_{L_p}\}$, these metrics are defined as,    
\begin{equation}
\label{eqn:dicescore}
    \text{DSC} = 2\frac{|G \cap P|}{|G| + |P|}
\end{equation}

\begin{equation}
\label{eqn:fpv}   
    \text{FPV} = v_p \sum_{l=1}^{L_p} |P_l| \delta(|P_l \cap G|) 
\end{equation}

\begin{equation}
\label{eqn:fnv}  
    \text{FNV} = v_g \sum_{l=1}^{L_g} |G_l| \delta(|G_l \cap P|) 
\end{equation}
where $\delta(x):= 1$ for $x=0$ and $\delta(x):= 0$ otherwise. $v_g$ and $v_p$  represent the voxel volumes (in ml) for ground truth and predicted mask, respectively (with $v_p = v_g$ since the predicted mask was resampled to the original ground truth coordinates). The submitted algorithms were ranked separately for each of the three metrics and the final ranking was determined based on the formula: $0.5 \times \text{rank}_\text{DSC} + 0.25  \times  \text{rank}_\text{FPV} + 0.25  \times \text{rank}_\text{FNV}$. The function definitions for these metrics were obtained from the challenge GitHub page and can be accessed via this \href{https://github.com/lab-midas/autoPET/blob/master/val_script.py}{link}.

\section{Results and discussion}
\label{sec:results}
\begin{figure}[H]
    \centering
    \includegraphics[width=0.9\textwidth]{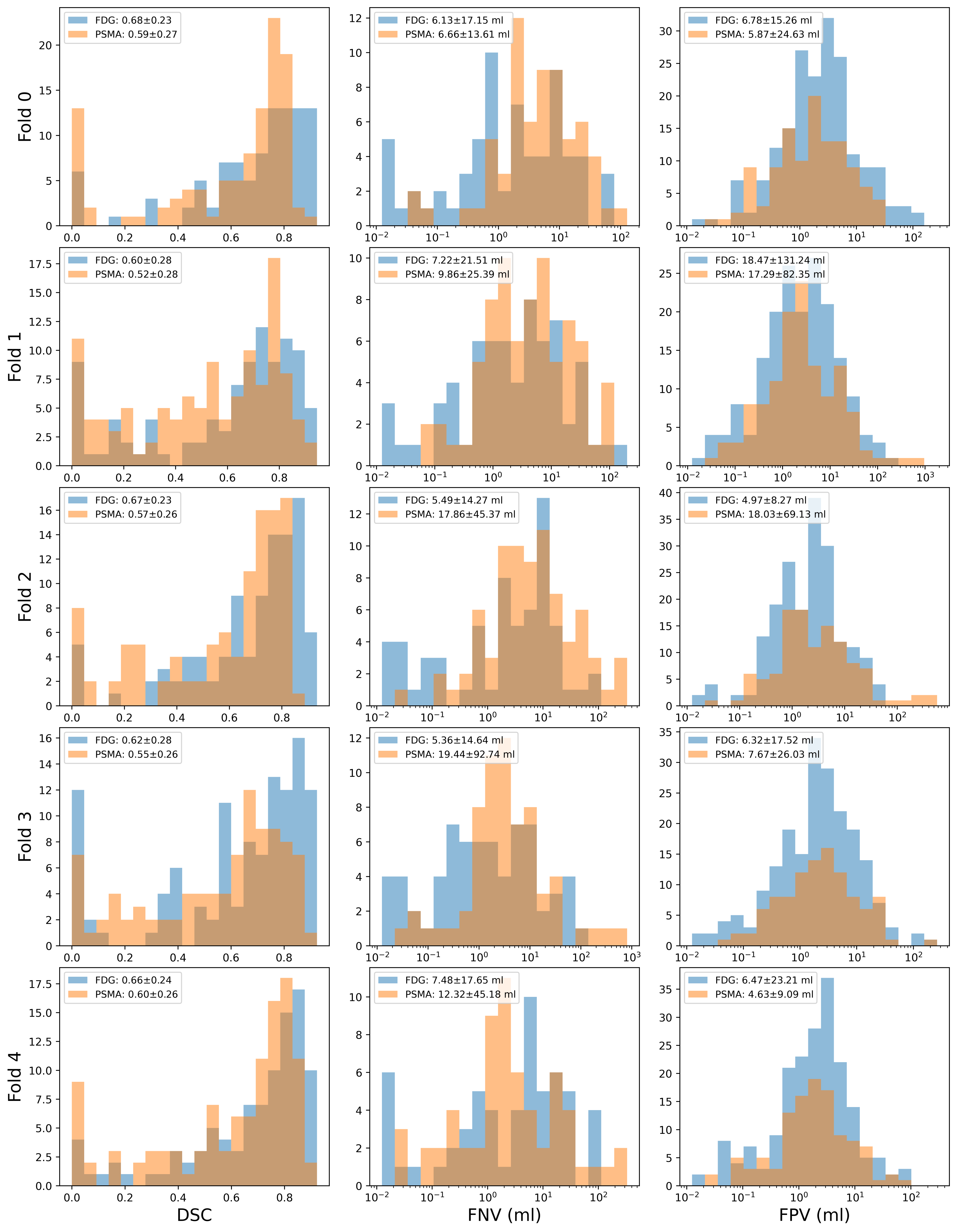}
    \caption{Distribution of DSC (left), FNV (middle) and FPV (right) from FDG (blue) and PSMA (orange) cases across each of the validation folds. The legend shows the mean values of the metrics over the two radiotracers. On an average, the networks performance on the FDG cases were superior than those on the PSMA cases across all folds, especially on the DSC and FPV metrics.}
    \label{fig:metrics_distr}
\end{figure}

\subsection{Five-fold cross validation results and the preliminary test phase}
We report the performance of the 5 models on the 5 validation folds in Table \ref{tab:validation_folds_results}. We report the mean values of the three metrics on the respective validation folds along with standard deviation. On the preliminary test set, the average ensemble obtained DSC, FNV, and FPV of 0.6687, 10.9522 ml, and 2.9684 ml, respectively, as shown in Table \ref{tab:prelim_results}.

\begin{table}[]
\centering
\caption{Metrics evaluated over five-fold cross-validation}
\label{tab:validation_folds_results}
\resizebox{0.6\columnwidth}{!}{%
\begin{tabular}{@{}cccc@{}}
\toprule
\multirow{2}{*}{\begin{tabular}[c]{@{}c@{}}Validation \\ fold (f)\end{tabular}} & \multicolumn{3}{c}{Mean $\pm$ std for metrics on fold $f$} \\ \cmidrule(l){2-4} 
  & DSC         & FNV (ml)      & FPV (ml)       \\ \midrule
0 & 0.6303 {\scriptsize $\pm$0.2563} & 6.4043 {\scriptsize $\pm$15.4433}  & 6.4421 {\scriptsize $\pm$19.3138}   \\
1 & 0.5534 {\scriptsize $\pm$0.2832} & 8.6698 {\scriptsize $\pm$23.8138}  & 17.9854 {\scriptsize $\pm$113.9501} \\
2 & 0.6211 {\scriptsize $\pm$0.2477} & 11.9128  {\scriptsize $\pm$34.7993} & 9.9178 {\scriptsize $\pm$43.5817}   \\
3 & 0.5848 {\scriptsize $\pm$0.2733} & 11.7539 {\scriptsize $\pm$63.9642} & 6.7691 {\scriptsize $\pm$20.7802}   \\
4 & 0.6302 {\scriptsize $\pm$0.2535} & 10.0395 {\scriptsize $\pm$35.1843} & 5.8035 {\scriptsize $\pm$19.4025}   \\ \bottomrule
\end{tabular}%
}
\end{table}

\begin{table}[]
\centering
\caption{Results on the preliminary test set}
\label{tab:prelim_results}
\resizebox{0.4\columnwidth}{!}{%
\begin{tabular}{@{}cccc@{}}
\toprule
Ensemble & DSC    & FNV (ml) & FPV (ml) \\ \midrule
Average  & 0.6687 & 10.9522  & 2.9684   \\ \bottomrule
\end{tabular}%
}
\end{table}

\subsection{Performance across FDG vs. PSMA cases}

\begin{figure}[h]
    \centering
    \includegraphics[width=0.8\textwidth]{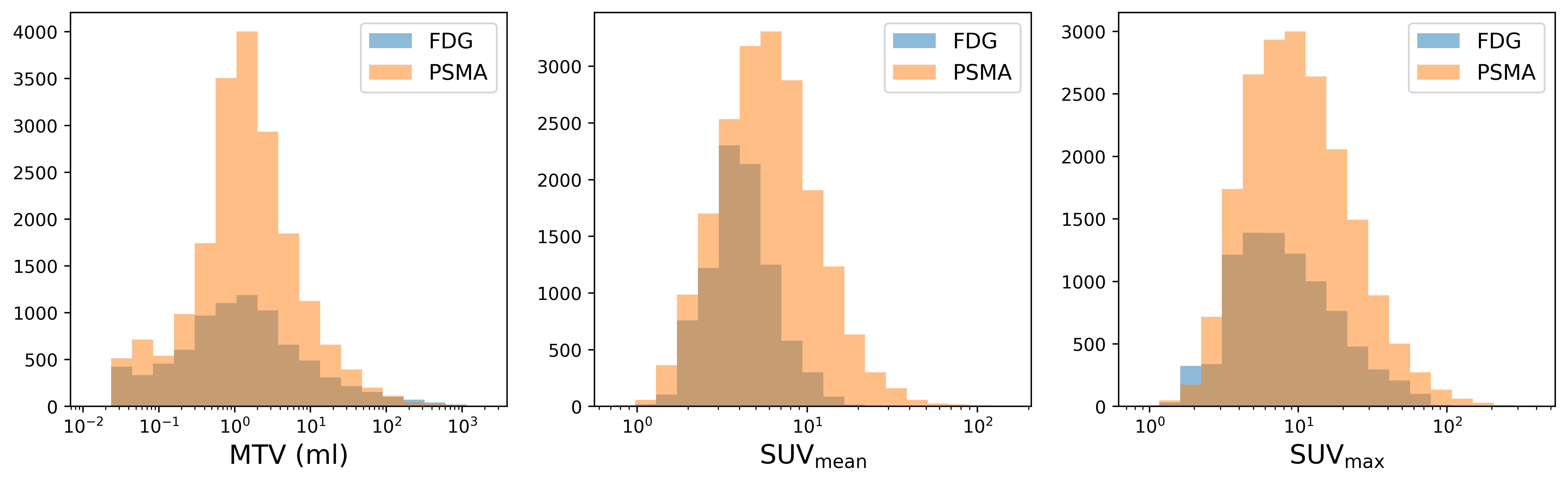}
    \caption{Ground truth lesion-level lesion measures such as lesion MTV (left), SUV$_\text{mean}$ (middle), and SUV$_{\text{max}}$ (right) across FDG (blue) and PSMA (orange) cases in the training set.}
    \label{fig:lesion_level_lesion_measures}
\end{figure}

\begin{figure}[htbp]
    \centering
    \includegraphics[width=0.8\textwidth]{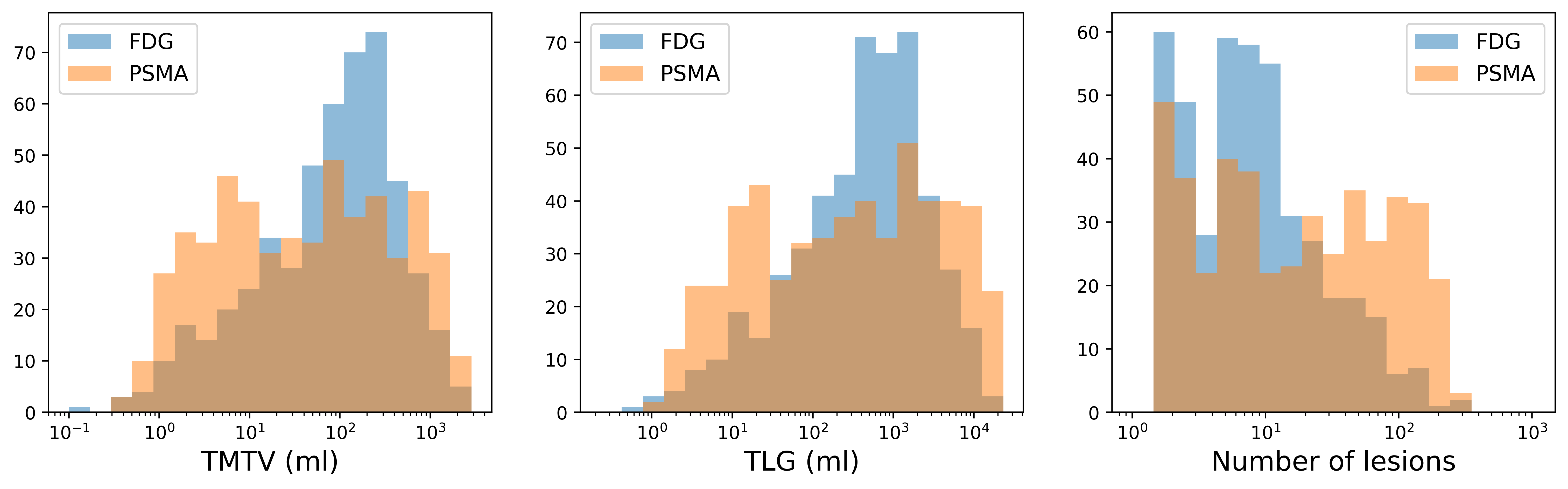}
    \caption{Ground truth patient-level lesion measures such as TMTV (left), TLG (middle), and number of lesions (right) across FDG (blue) and PSMA (orange) cases in the training set.}
    \label{fig:patient_level_lesion_measures}
\end{figure}

\begin{figure}[h]
    \centering
    \includegraphics[width=0.9\textwidth]{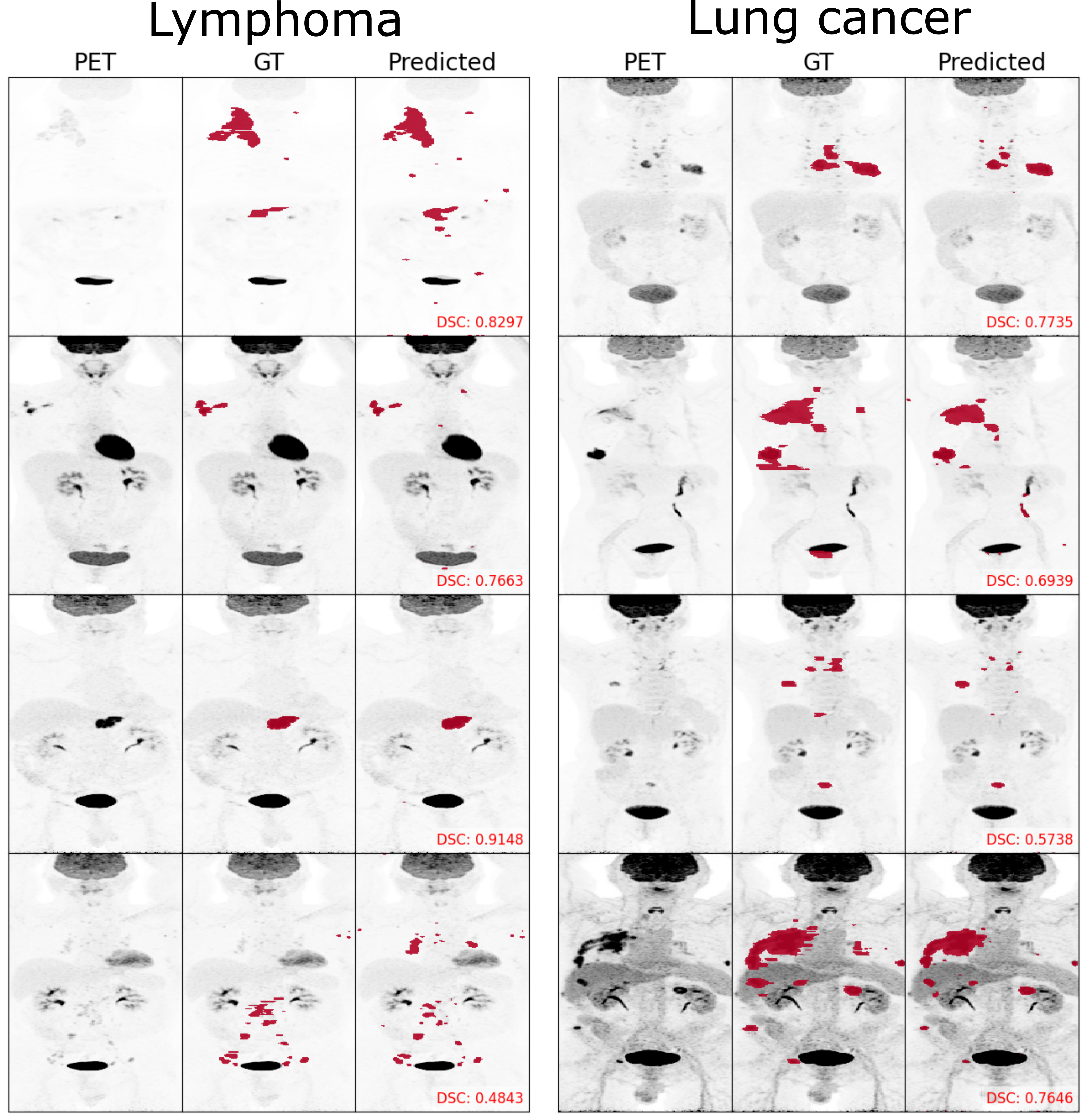}
    \caption{Some example cases showing the comparison between the ground truth and predicted lesion segmentation masks for four cases of lymphoma (left) and four cases of lung cancer (right).}
    \label{fig:lymphoma_lung_visual}
\end{figure}

\begin{figure}[h]
    \centering
    \includegraphics[width=0.9\textwidth]{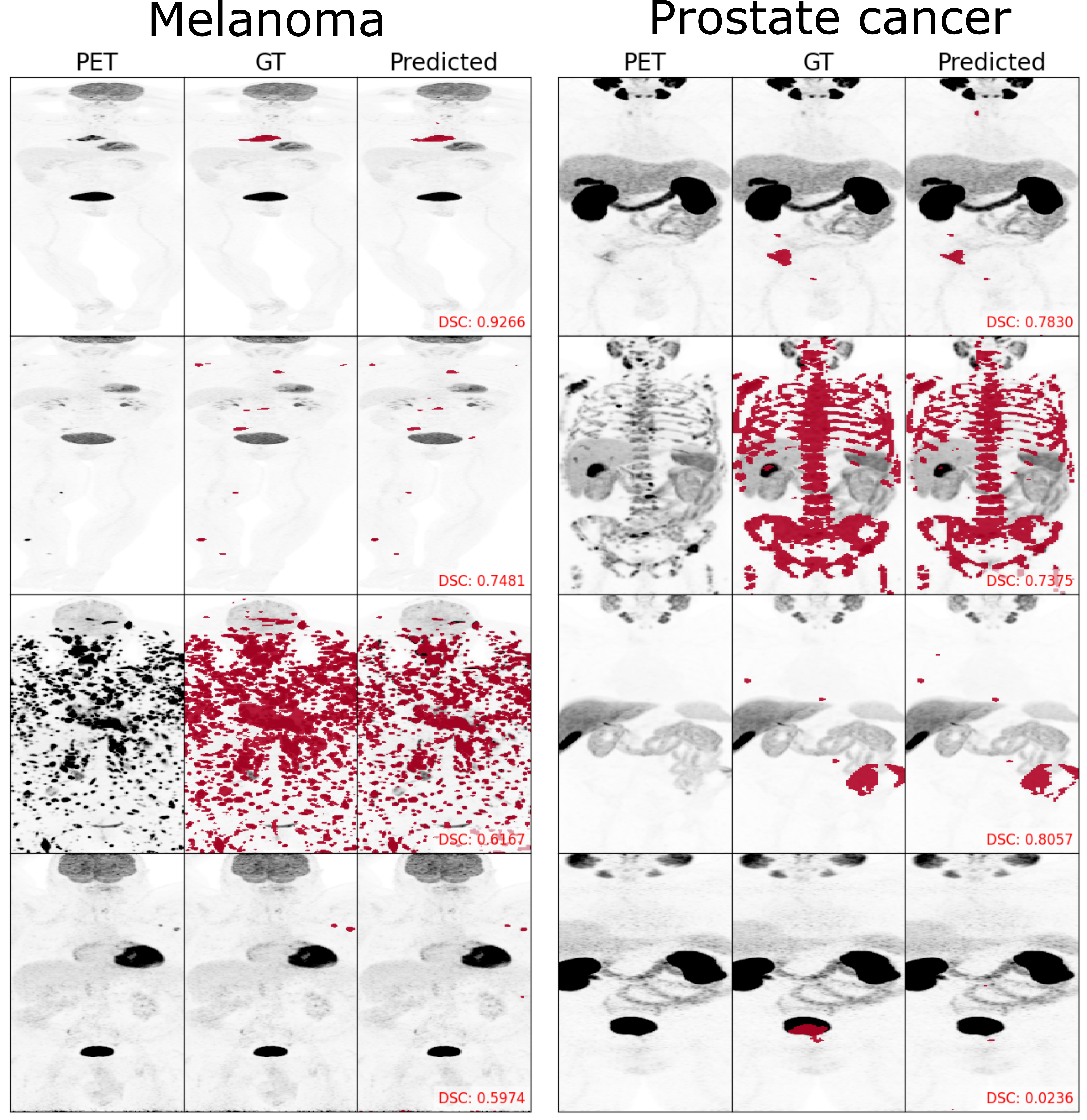}
    \caption{Some example cases showing the comparison between the ground truth and predicted lesion segmentation masks for four cases of melanoma (left) and four cases of prostate cancer (right).}
    \label{fig:melanoma_prostate_visual}
\end{figure}

Fig. \ref{fig:metrics_distr} shows the distribution of the 3 metrics for FDG and PSMA cases over all validation folds. We can see that, in general, the performance of the networks on FDG cases were superior on the FDG cases, as compared to the PSMA cases. This is especially true for the DSC metric where the FDG cases had mean DSCs of $0.68 \pm 0.23$, $0.60 \pm 0.28$, $0.67 \pm 0.23$, $0.62 \pm 0.28$, and $0.66 \pm 0.24$, while the PSMA cases had mean DSCs of $0.59 \pm 0.27$, $0.52 \pm 0.28$, $0.57 \pm 0.26$, $0.55 \pm 0.26$, and $0.60 \pm 0.26$ respectively over the five folds. The FDG cases also had superior values for FNV with means of $6.13 \pm 17.15$ ml, $7.22 \pm 21.51$ ml, $5.49 \pm 14.27$ ml, $5.36 \pm 14.64$ ml, and $7.48 \pm 17.65$ ml, while the PSMA cases had mean FNVs of $6.66 \pm 13.61$ ml, $9.86 \pm 25.39$ ml, $17.86 \pm 45.37$ ml, $19.44 \pm 92.74$ ml, and $12.32 \pm 45.18$ ml respectively over the five folds. Finally, the FDG cases had nearly comparable mean FPV values to the PSMA cases, with FDG having mean FPVs of $6.78 \pm 15.26$ ml, $18.47 \pm 131.24$ ml, $4.97 \pm 8.27$ ml, $6.32 \pm 17.52$ ml, and $6.47 \pm 23.21$ ml, while PSMA cases had mean FPV of $5.87 \pm 24.63$ ml, $17.29 \pm 82.35$ ml, $18.03 \pm 69.13$ ml, $7.67 \pm 26.03$ ml, and $4.63 \pm 9.09$ ml, respectively over the five folds. Some visualisations of the ground truth and predicted segmentation masks have been presented in Figs. \ref{fig:lymphoma_lung_visual} and \ref{fig:melanoma_prostate_visual}. \\

This trend can be attributed to the fact that a larger proportion of the training set consisted of FDG cases. Moreover, recent studies \cite{ahamed2023comprehensive, xu2023automatic} have shown that the performance of the segmentation networks on PET images are highly dependent on certain lesion characteristics or measures, and the network performs better on cases with higher values of these lesion measures. These measures include certain \textbf{lesion-level measures} such as (i) metabolic tumor volume (MTV), which represent the volume of individual tumor/lesion on PET images, (ii) lesion SUV$_\text{mean}$, which represents the mean SUV value over all the voxels in a lesion, (iii) lesion SUV$_\text{max}$, which is the maximum SUV over all the voxels in a lesion, or certain \textbf{patient-level lesion measures} such as, (iv) total metabolic tumor volume (TMTV), which is the sum of the MTVs of all the tumors inside a patient, (v) total lesion glycolysis (TLG), which is the sum of the product of lesion SUV$_\text{mean}$ and lesion MTV over all lesions in a patient, and (vi) total number of lesions in a patient. The ground truth distribution for lesions-level and patient-level lesion measures have been shown in Figs. \ref{fig:lesion_level_lesion_measures} and \ref{fig:patient_level_lesion_measures}, respectively. From our analysis of the ground truth training data, we observe that although the number of FDG cases is larger than that of PSMA, there were a total of 8,781 FDG lesions while 19,377 PSMA lesions in the training set. From the left figures of Figs. \ref{fig:lesion_level_lesion_measures} and \ref{fig:patient_level_lesion_measures}, we can see that although the there are more PSMA lesions, the overall TMTV of PSMA cases is skewed to the left, meaning that a considerable fraction of PSMA lesions are smaller in size. These smaller lesions were particularly challenging to segment. Moreover, for an image contain a large number of lesions (like a typical PSMA case), the network tended to segment the brightest (higher SUV$_\text{mean}$ or TLG) lesions while ignoring the rest, leading to lower network performance on the PSMA cases.

\section{Conclusion}
\label{conclusion}
In this study, we developed a deep Residual UNet model, trained using a 5-fold cross-validation approach, where we optimized a Generalized Dice Focal loss objective function to segment lesions from a dataset consisting of FDG and PSMA radiotracers-based PET/CT images. As a direction for future research, we plan to enhance the objective function by incorporating terms for False Positive Volume (FPV) and False Negative Volume (FNV) to more effectively penalize predictions with high FPV or FNV, beyond the use of a Dice-based loss alone.

\section{Acknowledgment}
I would like to acknowledge the support of my friend H. Fan for providing me with a GPU for this work over several months.

\nopagebreak


\begin{thebibliography}{99}

\bibitem{pet_importance}
S. F. Barrington and Michel Meignan,
\emph{Time to Prepare for Risk Adaptation in Lymphoma by Standardizing Measurement of Metabolic Tumor Burden},
\emph{Journal of Nuclear Medicine},
\textbf{60}(8), 1096-1102, 2019.

\bibitem{tmtv}
Tarec Christoffer El-Galaly, Diego Villa, Chan Yoon Cheah, Lars C. Gormsen, et al.,
\emph{Pre-treatment total metabolic tumour volumes in lymphoma: Does quantity matter?},
\emph{British Journal of Haematology},
\textbf{197}(2), 139-155, 2022.

\bibitem{suvmean_tmtv_tlg}
Kursat Okuyucu, Sukru Ozaydin, Engin Alagoz, Gokhan Ozgur, et al.,
\emph{Prognosis estimation under the light of metabolic tumor parameters on initial FDG-PET/CT in patients with primary extranodal lymphoma},
\emph{Radiol. Oncol.},
\textbf{50}(4), 360-369, 2016.

\bibitem{FOSTER201476}
Brent Foster, Ulas Bagci, Awais Mansoor, Ziyue Xu, Daniel J. Mollura,
\emph{A review on segmentation of positron emission tomography images},
\emph{Computers in Biology and Medicine},
\textbf{50}, 76-96, 2014.

\bibitem{Wang2021}
Shanshan Wang, Cheng Li, Rongpin Wang, Zaiyi Liu, Meiyun Wang, Hongna Tan, Yaping Wu, Xinfeng Liu, Hui Sun, Rui Yang, Xin Liu, Jie Chen, Huihui Zhou, Ismail Ben Ayed, Hairong Zheng,
\emph{Annotation-efficient deep learning for automatic medical image segmentation},
\emph{Nature Communications},
\textbf{12}(1), 5915, 2021.

\bibitem{Tang2019}
Hao Tang, Xuming Chen, Yang Liu, Zhipeng Lu, Junhua You, Mingzhou Yang, Shengyu Yao, Guoqi Zhao, Yi Xu, Tingfeng Chen, Yong Liu, Xiaohui Xie,
\emph{Clinically applicable deep learning framework for organs at risk delineation in CT images},
\emph{Nature Machine Intelligence},
\textbf{1}(10), 480-491, 2019.

\bibitem{Hollon2020}
Todd C. Hollon, et al.,
\emph{Near real-time intraoperative brain tumor diagnosis using stimulated Raman histology and deep neural networks},
\emph{Nature Medicine},
\textbf{26}(1), 52-58, 2020.

\bibitem{Gatidis2022}
Sergios Gatidis, Tobias Hepp, Marcel Früh, Christian La Fougère, Konstantin Nikolaou, Christina Pfannenberg, Bernhard Schölkopf, Thomas Küstner, Clemens Cyran, Daniel Rubin,
\emph{A whole-body FDG-PET/CT Dataset with manually annotated Tumor Lesions},
\emph{Scientific Data},
\textbf{9}(1), 601, 2022.

\bibitem{autopet2022_paper}
Sergios Gatidis, Marcel Früh, Matthias Fabritius, et al,
\emph{The autoPET challenge: Towards fully automated lesion segmentation in oncologic PET/CT imaging},
\emph{Research Square}, 2023.

\bibitem{resunet}
Eric Kerfoot, James Clough, Ilkay Oksuz, Jack Lee, Andrew P. King, Julia A. Schnabel,
\emph{Left-Ventricle Quantification Using Residual U-Net},
\emph{Statistical Atlases and Computational Models of the Heart. Atrial Segmentation and LV Quantification Challenges},
Springer International Publishing, 2018.

\bibitem{monai_paper}
M. Jorge Cardoso, Wenqi Li, Richard Brown, et al.,
\emph{MONAI: An open-source framework for deep learning in healthcare},
arXiv preprint arXiv:2211.02701, 2022.

\bibitem{gendiceloss}
Carole H. Sudre, Wenqi Li, Tom Vercauteren, Sebastien Ourselin, M. Jorge Cardoso,
\emph{Generalised Dice Overlap as a Deep Learning Loss Function for Highly Unbalanced Segmentations},
\emph{Deep Learning in Medical Image Analysis and Multimodal Learning for Clinical Decision Support},
Springer International Publishing, 2017.

\bibitem{focalloss}
Tsung-Yi Lin, Priya Goyal, Ross Girshick, Kaiming He, Piotr Dollár,
\emph{Focal Loss for Dense Object Detection},
arXiv preprint arXiv:1708.02002, 2018.

\bibitem{ahamed2023comprehensive}
Ahamed, S. and Xu, Y., et al,
\emph{Comprehensive Evaluation and Insights into the Use of Deep Neural Networks to Detect and Quantify Lymphoma Lesions in PET/CT Images},
arXiv preprint arXiv:2311.09614.

\bibitem{ahamed2023generalized}
Ahamed, S. and Rahmim, A.,
\emph{Generalized Dice Focal Loss trained 3D Residual UNet for Automated Lesion Segmentation in Whole-Body FDG PET/CT Images},
arXiv preprint arXiv:2309.13553.

\bibitem{xu2023automatic}
Xu, Y., et al,
\emph{Automatic segmentation of prostate cancer metastases in PSMA PET/CT images using deep neural networks with weighted batch-wise dice loss},
Computers in Biology and Medicine, 2023.



\end{thebibliography}
\end{document}